\documentclass{article} 
\usepackage{iclr2026_conference,times}

\usepackage{amsmath,amsfonts,bm}

\def\eqref#1{equation~\ref{#1}}

\def\1{\bm{1}}

\DeclareMathAlphabet{\mathsfit}{\encodingdefault}{\sfdefault}{m}{sl}
\SetMathAlphabet{\mathsfit}{bold}{\encodingdefault}{\sfdefault}{bx}{n}

\def\gC{{\mathcal{C}}}
\def\gD{{\mathcal{D}}}

\def\gI{{\mathcal{I}}}

\def\gK{{\mathcal{K}}}

\def\gR{{\mathcal{R}}}
\def\gS{{\mathcal{S}}}

\def\gX{{\mathcal{X}}}
\def\gY{{\mathcal{Y}}}

\usepackage{multirow}
\usepackage{hyperref}
\usepackage{booktabs} 
\usepackage{url}
\usepackage{pdflscape}
\usepackage{subfigure}
\usepackage{graphicx}
\usepackage{makecell}
\usepackage{algorithm}
\usepackage{algorithmic}
\usepackage{amsmath}
\usepackage{wrapfig}
\usepackage{svg}
\usepackage{listings}
\usepackage{multirow}
\usepackage{tabularx}
\usepackage[table]{xcolor}
\usepackage{array}
\usepackage[most]{tcolorbox}
\usepackage{xcolor}
 \usepackage{amssymb}
 
\definecolor{c1}{RGB}{239,118,122}
\definecolor{c2}{RGB}{69,105,144}
\definecolor{c3}{RGB}{72,192,170}
\definecolor{c4}{RGB}{179,149,189}

\title{Model-Document Protocol for AI Search}

\author{%
  Hongjin Qian, Zheng Liu\thanks{Corresponding Author} \\
  BAAI\\
  \texttt{\{chienqhj, zhengliu1026\}@gmail.com} 
}

\iclrfinalcopy 
\begin{document}

\maketitle

\begin{abstract}
AI search depends on linking large language models (LLMs) with vast external knowledge sources. Yet web pages, PDF files, and other raw documents are not inherently LLM-ready: they are long, noisy, and unstructured. Conventional retrieval methods treat these documents as verbatim text and return raw passages, leaving the burden of fragment assembly and contextual reasoning to the LLM. This gap underscores the need for a new retrieval paradigm that redefines how models interact with documents.

We introduce the Model–Document Protocol (MDP), a general framework that formalizes how raw text is bridged to LLMs through consumable knowledge representations. Rather than treating retrieval as passage fetching, MDP defines multiple pathways that transform unstructured documents into task-specific, LLM-ready inputs. These include agentic reasoning, which curates raw evidence into coherent context; memory grounding, which accumulates reusable notes to enrich reasoning; and structured leveraging, which encodes documents into formal representations such as graphs or KV caches. All three pathways share the same goal: ensuring that what reaches the LLM is not raw fragments but compact, structured knowledge directly consumable for reasoning.

As an instantiation, we present MDP-Agent, which realizes the protocol through an agentic process: constructing document-level gist memories for global coverage, performing diffusion-based exploration with vertical exploitation to uncover layered dependencies, and applying map–reduce style synthesis to integrate large-scale evidence into compact yet sufficient context. Experiments on information-seeking benchmarks demonstrate that MDP-Agent outperforms baselines, validating both the soundness of the MDP framework and the effectiveness of its agentic instantiation. Our codes are available in this \href{https://github.com/qhjqhj00/MDP-Agent}{\textit{repository}}.
\end{abstract}

\begin{figure}[h]
    \centering
    \includegraphics[width=0.8\linewidth]{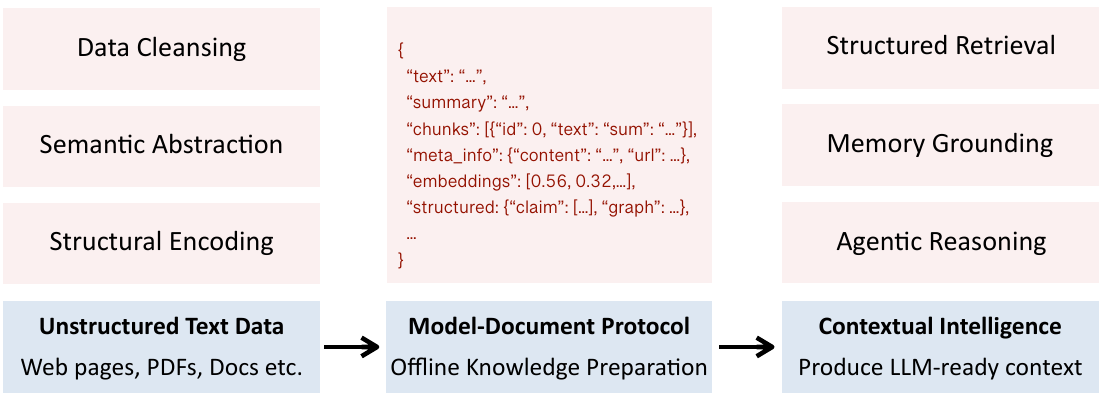}
    \caption{The Model–Document Protocol (MDP) provides a standard protocol for bridging unstructured data to LLMs. Raw documents are first processed through basic cleaning and segmentation, followed by semantic abstraction that captures summaries, claims, and topics. These signals are then transformed into structured formats such as graphs or KV caches. Building on this foundation, MDP enables contextual intelligence for LLMs through multiple pathways, including constructing LLM-ready context via agentic information discovery, enhancing reasoning with memory grounding, and directly leveraging pre-encoded structures such as KV caches or graphs.}
    \label{fig:mdp}
\end{figure}

\section{Introduction}

Large language models (LLMs) have recently shown impressive capabilities in information-seeking tasks, producing coherent responses to queries ranging from factual lookup to multi-step reasoning~\citep{chatgpt,gemini25,deepseekr1}. However, their knowledge remains fixed to the pretraining corpus, leaving critical gaps in coverage and timeliness. As a result, LLMs struggle on knowledge-intensive tasks that demand specialized expertise or up-to-date information~\citep{zhao2024surveylargelanguagemodels,hallucination}.

A common solution is to augment LLMs with external sources such as the web or local knowledge bases. The prevailing paradigm, retrieval-augmented generation (RAG), grounds responses by injecting retrieved passages into the model’s context~\citep{rag20,gao2024retrievalaugmented}. While effective for many factual queries, RAG often falters on complex tasks where evidence is interdependent and requires multi-step reasoning~\citep{zhao2024retrievalaugmentedgenerationrag}. Tool-integrated reasoning methods extend this idea by interleaving retrieval with reasoning, enabling agents to iteratively refine queries, call tools, and synthesize evidence~\citep{searchr1,li2025webthinker}. Yet their reliance on in-context evidence processing makes them inefficient and difficult to scale as context length grows.

Beyond these methodological limits lies a deeper obstacle: when grounded in real-world data, LLMs inevitably face what we term \emph{Data Chaos}. Knowledge in practice resides in long-form, heterogeneous, unstructured, noisy, and redundant sources such as web pages and PDF files~\citep{qian2023webbrain,zhu2024largelanguagemodelsinformation}. Relevant information is sparsely scattered across massive volumes of text, yielding a low signal-to-noise ratio. As context windows saturate, LLMs struggle to isolate and assemble the crucial evidence required to bridge task-specific knowledge gaps. Effective information seeking therefore demands more than retrieving fragments—it requires reorganizing raw documents into coherent, structured evidence that LLMs can reliably exploit~\citep{zhang2024agentic}.

In this paper, we introduce the Model–Document Protocol (MDP), a general framework that redefines retrieval as the transformation of raw, unstructured data into compact, task-specific knowledge directly consumable by LLMs. Rather than treating retrieval as mere passage fetching, MDP specifies how documents should be abstracted, explored, and synthesized so that fragmented, noisy sources are reorganized into coherent evidence that bridges task-specific knowledge gaps. Crucially, the protocol allows multiple pathways to contextual intelligence, including agentic reasoning that curates LLM-ready context, memory grounding that accumulates reusable notes, and structured leveraging that encodes documents into formal representations such as graphs or KV caches.

\begin{figure}[t]
    \centering
    \includegraphics[width=\linewidth]{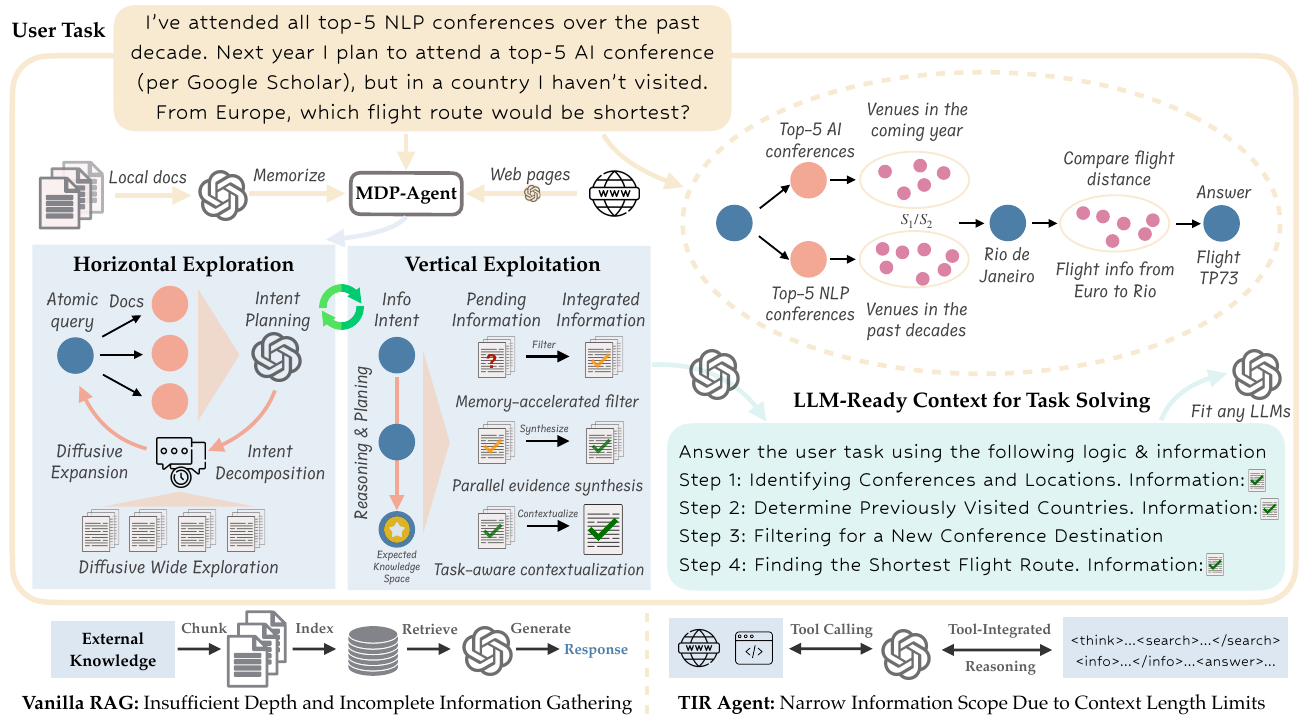}
    \caption{Illustration of a complex information-seeking task, where the answer depends on satisfying multiple conditions through horizontal exploration and vertical exploitation.  MDP-Agent addresses this agentically by formulating intents, decomposing them into atomic queries, and expanding coverage via diffusion to gather raw documents. Resolved intents advance iteratively to the next, with documents processed in parallel and synthesized through a map–reduce procedure into subspace knowledge, which is then transformed into an LLM-ready context.
}
    \label{fig:model}
\end{figure}

As a concrete instantiation, we present MDP-Agent (Figure~\ref{fig:model}). At the corpus level, MDP-Agent organizes raw sources into a structured warehouse: each document is paired with a gist memory that abstracts its global theme and structure, and indexed through a hybrid scheme combining dense embeddings with sparse text, enabling both global coverage and fine-grained access. At the task level, given a complex query, MDP-Agent plans information intents, decomposes them into atomic sub-queries, and retrieves the corresponding atomic knowledge units. Coverage is expanded through diffusion-based horizontal exploration, while layered dependencies are resolved through vertical exploitation.
To maintain scalability, MDP-Agent leverages memory-accelerated filtering via gist memories to pre-select relevant documents and applies map–reduce–style parallel synthesis to integrate large-scale evidence outside the main reasoning loop. The resulting evidence is then composed into a structured knowledge chain and reformulated into a compact, LLM-ready context. While MDP-Agent exemplifies one instantiation of MDP, the protocol itself is more general and admits diverse implementations that operationalize its principles of abstraction, exploration, and synthesis in different ways.

To evaluate the effectiveness of our approach, we conduct extensive experiments on challenging information-seeking benchmarks. Results show that our method consistently outperforms strong baselines.
Our contributions are threefold:
(1)~We identify the fundamental mismatch between conventional retrieval methods, which return raw text passages, and the way LLMs consume knowledge, which requires structured and task-specific context. To address this gap, we introduce the Model–Document Protocol (MDP), a general framework that formalizes retrieval as the structured transformation of raw documents into compact knowledge consumable by LLMs.
(2)~We instantiate this protocol with MDP-Agent, which integrates document-level gist memory with an agentic decision process. It performs diffusion-based horizontal exploration and vertical exploitation, coupled with map–reduce style synthesis, to construct minimal yet sufficient knowledge spaces in an LLM-ready form.
(3)~We provide extensive empirical validation across multiple challenging benchmarks, demonstrating that MDP-Agent effectively connects large-scale external knowledge with LLM reasoning, offering a scalable and general solution for enhancing standalone LLMs.

\section{Model-Document Protocol: MDP}

\subsection{Preliminary}
\paragraph{Complex Information-Seeking Task.} 
Solving a task with an LLM can be formalized as \(\gY = \Theta(\gX \mid \gK)\), where $\Theta$ denotes the model’s generative function and \(\gK\) captures the knowledge required to bridge the gap between input and output. In this view, producing the correct answer amounts to filling a \emph{knowledge gap} that separates \(\gX\) from \(\gY\). When the task is simple fact-based or commonsense in nature, the gap is typically small and can often be resolved by the model’s pretrained knowledge or a single retrieval step.
In contrast, complex tasks create a much larger and more intricate gap. Recovering the expected knowledge space \(\gK\) in such cases is challenging, requiring multi-layered exploration and iterative decision-making in which evidence is progressively gathered, refined, and integrated until the gap is sufficiently closed to yield a reliable answer.

For \emph{complex information-seeking tasks}, solving the problem typically unfolds as a multi-step reasoning process, where the required knowledge emerges in stages rather than all at once. In this setting, the knowledge gap \(\gK\) can be formally represented as a sequential \emph{knowledge chain}:
\begin{equation}
    \gK = (\gK_1 \rightarrow \gK_2 \rightarrow \cdots \rightarrow \gK_t),
    \label{eq:def}
\end{equation}
where \(\gK_i\) denotes the crucial knowledge required at the \(i\)-th reasoning step, and the arrow \(\rightarrow\) indicates the sequential dependency among steps. Each intermediate knowledge space \(\gK_i\) is itself formed by combining multiple \emph{atomic knowledge spaces}:
\begin{equation}
    \gK_i = \gS_{i,1} \cap \gS_{i,2} \cap \cdots \cap \gS_{i,n_i},
\end{equation}
where each \(\gS_{i,j}\) represents a minimal unit of knowledge that can be directly retrieved through a single query $q$, typically consisting of a set of relevant documents \(\{D\}\)\footnote{Here, a ``document'' is used in a broad sense and may refer to a web page, a PDF file, or a full text piece.}.

This formulation emphasizes two complementary dimensions of reasoning. \emph{Depth} arises from the sequential composition of the knowledge chain, while \emph{breadth} comes from the conjunction of multiple atomic knowledge spaces within each step. Special cases follow naturally: when \(t=1, n_i=1\), the task reduces to a single-hop factual query; when \(t>1\) but each \(n_i=1\), it corresponds to simple multi-hop reasoning over independent facts.  

From an information-theoretic perspective, the \emph{knowledge gap} \(\gK\) quantifies the additional information required to determine the correct answer \(\gY\) given input \(\gX\). Solving a complex task can thus be seen as an iterative reduction of conditional entropy,
\begin{align}
      H(\gY \mid \gX) \;>\; H(\gY \mid \gX,\gK_1) \;>\; \cdots \;>\; H(\gY \mid \gX,\gK_1,\dots,\gK_t)=0.
\end{align}
Here, \emph{breadth} aggregates multiple atomic sources that jointly constrain uncertainty, while \emph{depth} reflects the sequential dependencies through which uncertainty is progressively eliminated. In this view, the \emph{knowledge chain} functions as an information channel that transmits the missing bits required to close the gap between input and answer.

\paragraph{Data Chaos: The Bottleneck for LLM Context.}
We refer to \emph{Data Chaos} as the state in which essential knowledge is entangled within vast amounts of unstructured, redundant, and noisy data that an LLM cannot directly exploit. Suppose the universal knowledge space is denoted by \(\gS\). In principle, one could hope to isolate a \emph{minimal but sufficient} subset \(\gK^* \subset \gS\) that contains exactly the information required for producing the answer \(\gY\) and nothing more. Such a representation would be ideal, as it would minimize entropy and maximize the signal-to-noise ratio of the input context.  

In practice, however, retrieval for complex tasks produces large volumes of raw text (e.g., hundreds of web pages or thousands of PDF documents) that remain far from this ideal. The data are dominated by formatting artifacts, boilerplate language, and irrelevant content, with useful knowledge sparsely embedded within. Feeding such raw collections into an LLM is both inefficient and ineffective: the context window is quickly saturated, and the high entropy of the input obscures the information truly relevant to the task. In this sense, retrieved data in their raw form are not \emph{LLM-ready}, but rather exemplify the disorder of \emph{Data Chaos}.

\subsection{Formal Definition of the Model–Document Protocol}

\paragraph{Motivation.}
The goal of the \textbf{Model–Document Protocol (MDP)} is to formalize how unstructured, high-entropy data are progressively transformed into compact, structured, and LLM-consumable knowledge representations.  
Conventional retrieval methods provide no such protocol: they retrieve raw text fragments and leave the reasoning model $\Theta$ to perform fragment assembly, relevance estimation, and contextual synthesis on its own.  
MDP introduces a principled interface between raw data and reasoning models, defining both \emph{what} should be transferred and \emph{how} it should be structured.  

\paragraph{Formalization.}
Let $\mathbb{D}$ denote the raw document corpus, composed of unstructured data sources such as web pages, PDFs, and full-text records:
\begin{equation}
    \mathbb{D} = \{ D_1, D_2, \ldots, D_N \}, \quad D_i \in \text{RawText}.
\end{equation}
A retrieval process based purely on keyword or embedding similarity yields a raw subset $\gR \subset \mathbb{D}$ that remains dominated by redundant and noisy information.
MDP defines a multi-stage transformation function $\Phi$, which converts the retrieved collection $\gR$ into a structured knowledge representation $\gK_\text{MDP}$ that is directly consumable by the model $\Theta$:
\begin{equation}
    \gK_\text{MDP} = \Phi(\gR; \Psi),
\end{equation}
where $\Psi$ represents the protocol specification—a set of transformation pathways and policies that govern how evidence is selected, abstracted, and represented.
Formally, $\Phi$ is decomposed into three complementary pathways:
\begin{equation}
    \Phi = \Phi_\text{agentic} \circ \Phi_\text{memory} \circ \Phi_\text{structured},
\end{equation}
corresponding respectively to:
\begin{itemize}
    \item \textbf{Agentic Reasoning Pathway} $\Phi_\text{agentic}$: performs iterative reasoning and evidence curation, assembling fragments from $\gR$ into coherent task-relevant context.
    \item \textbf{Memory Grounding Pathway} $\Phi_\text{memory}$: accumulates persistent, reusable notes extracted from prior interactions or retrieved data, enriching $\Phi_\text{agentic}$ with historical knowledge.
    \item \textbf{Structured Leveraging Pathway} $\Phi_\text{structured}$: encodes curated knowledge into structured forms such as key–value caches, relational graphs, or symbolic schemas, enhancing interpretability and reusability.
\end{itemize}

\paragraph{Protocol Semantics.}
Given an input query $\gX$, the LLM’s reasoning process under MDP can be expressed as:
\begin{equation}
    \gY = \Theta(\gX \mid \Phi(\gR; \Psi)),
\end{equation}
where $\Phi(\gR; \Psi)$ ensures that the conditioning knowledge is compact and structured, thereby minimizing contextual entropy:
\begin{equation}
    H(\Phi(\gR; \Psi)) \ll H(\gR).
\end{equation}
This entropy reduction encapsulates the core philosophy of MDP: transforming \emph{data chaos} into \emph{knowledge order}.  
The resulting $\gK_\text{MDP}$ serves as the bridge between raw documents and reasoning-ready inputs, establishing a unified interface through which diverse document forms can be processed, summarized, and consumed by LLMs.

\paragraph{Interpretation.}
In essence, MDP defines a mapping from raw document space to structured knowledge space:
\begin{equation}
    \text{MDP: } \mathbb{D} \rightarrow \gK_\text{MDP} \rightarrow \Theta,
\end{equation}
analogous to a communication protocol that guarantees the fidelity and efficiency of information exchange between humans and machines.  
Different instantiations of MDP implement this protocol through specific strategies (e.g., hierarchical diffusion, memory-based synthesis, or graph-structured reasoning), while adhering to the same formal semantics defined above.

\section{An Instantiation: MDP-Agent}

To tackle the challenge of Data Chaos, we propose  MDP-Agent, a initial instantiation of MDP that iteratively explores and exploits external knowledge, distilling high-entropy and noisy retrieval results into a minimal yet sufficient knowledge space, which is then transformed into task-specific, LLM-ready context for enabling contextual intelligence.  MDP-Agent operates in two stages: \emph{data indexing} and \emph{agentic knowledge discovery}, which we detail in the following sections.

\subsection{Data Indexing with Gist Memory}
\label{sec:gist}
Indexing real-world text data is challenging: document lengths vary from short news snippets to full academic papers, and structures are highly diverse, as in web pages with templates, metadata, and mixed formats. Direct applying dense retrieval to such raw text struggles under these conditions: for very long documents, it lacks global awareness because the encoder can only process limited windows; for structurally complex documents, it fails to capture implicit layout or organizational cues that are not explicitly encoded in raw text.  

To overcome these limitations,  MDP-Agent introduces an intermediate representation that makes implicit global and structural information explicit. For each document \(D\), a lightweight long-context model produces a textual abstraction \(\gD\) that verbalizes the document’s high-level topics and structural cues while omitting details. For example, a journal issue page may be represented by its title, scope, and the categories of included articles, without enumerating the individual titles. This abstraction parallels the way humans process long texts: after reading, we tend to retain only a gist-level memory that preserves the overall theme and structure while discarding fine-grained details. We therefore refer to \(\gD\) as a form of \emph{gist memory}. Unlike standard dense embeddings, which are typically derived from limited text windows and thus capture primarily local semantics, gist memory encodes a document’s global theme and structure, such as topical hierarchy and organizational flow. This richer abstraction allows retrieval methods not only to locate broadly relevant documents but also to filter and prioritize them more effectively, accessing cues that would otherwise be overlooked by dense representations alone.

Indexing then proceeds in a hybrid manner. Each gist representation \(\gD\) is encoded into a dense vector \(\mathbf{z}_D \in \mathbb{R}^d\) to capture global semantics, while the raw document \(D\) is indexed using a sparse scheme to retain fine-grained evidence. Given a query \(q\), relevance is computed as:
\begin{align}
    \text{Rel}(q, D) = \alpha \cdot \text{sim}_{\text{dense}}(q, \mathbf{z}_D) 
    + (1-\alpha) \cdot \text{sim}_{\text{sparse}}(q, D),
    \label{eq:ret}
\end{align}
where \(\alpha \in [0,1]\) balances semantic coherence from dense matching with detail sensitivity from sparse matching. This \emph{gist-memory based hybrid index} ensures that retrieval remains both globally  MDP-Agent and locally precise, enabling effective navigation of heterogeneous knowledge sources.

\subsection{Agentic Knowledge Discovery}
By the definition of Eq.~(\ref{eq:def}), the expected knowledge space \(\gK\) for a complex task cannot be obtained in a single step but must be assembled hierarchically. Specifically, \(\gK\) is composed of a sequence of subspaces \(\{\gK_1, \gK_2, \ldots, \gK_N\}\), where each \(\gK_i\) captures the evidence needed to resolve one stage of reasoning. In turn, every \(\gK_i\) is constructed from a collection of atomic knowledge spaces \(\{\gS_{i,1}, \gS_{i,2}, \ldots, \gS_{i,m_i}\}\), each of which can be retrieved by issuing a single query $q$. 

 MDP-Agent constructs this knowledge in an agentic manner. Given a task \(\gX\), the system reasons over it to issue an initial information intent \(I_1\), identifies the underlying knowledge gaps, and decomposes \(I_1\) into atomic sub-queries:
\begin{align}
    I_1 \;\mapsto\; \{q_{1,1}, q_{1,2}, \ldots, q_{1,m_1}\}.
\end{align}
Each sub-query \(q_{1,j}\) corresponds to a concrete retrieval action defined in Eq.~(\ref{eq:ret}), yielding an atomic knowledge space \(\gS_{1,j}\). Collectively, these atomic spaces form the subspace \(\gK_1\). Once \(\gK_1\) provides sufficient evidence to resolve \(I_1\), the process advances to the next intent \(I_2\), producing \(\gK_2\) in the same manner. This sequential procedure continues until all subspaces are constructed, yielding an approximation of the expected knowledge space \(\gK\) required to solve the task.
This process can be expressed recursively as:
\begin{align}
\gY = \Theta_N\Big(\ldots, \Theta_2(\Theta_1(\gX \mid I_1) \cup \gK_1) \cup \gK_2, \ldots \cup \gK_{N-1}\Big),
\label{eq:reasoning}
\end{align}
where \(I_t\) denotes the information intent at step \(t\), \(\gK_t\) is the corresponding subspace constructed from atomic spaces, and \(\Theta_t\) is the reasoning operation that advances once \(I_t\) is resolved. 

Building on this general definition, we instantiate  MDP-Agent with three core mechanisms: \emph{Diffusive Wide Exploration} for knowledge coverage, \emph{Memory-Guided Parallel Synthesis} for processing efficiency, and \emph{Task- MDP-Agent Contextualization} for synthesizing LLM-ready context. These components will be introduced in detail below.

\paragraph{Diffusive Wide Exploration.}

A central challenge in constructing a subspace \(\gK_i\) for a given intent \(I_i\) lies in \emph{intent alignment}: the description of \(I_i\) may be biased or incomplete, so its initial sub-queries may fail to cover the expected subspace. To mitigate this,  MDP-Agent employs a \emph{Diffusion Search} strategy designed to maximize the coverage of intent-relevant knowledge. After executing the initial queries and obtaining atomic spaces, the agent evaluates whether the accumulated evidence suffices for \(I_i\). If not, it expands the search frontier by generating additional queries conditioned on past results, thereby progressively enlarging the retrieved knowledge space:
\begin{align}
    \{q_{i,1}, \ldots, q_{i,m_i}\} 
    \;\;\mapsto\;\; \{\gS_{i,1}, \ldots, \gS_{i,m_i}\} 
    \;\;\mapsto\;\;
    \begin{cases}
        \gK_i, & \text{if sufficient}, \\
        \{q_{i,m_i+1}, \ldots\}, & \text{otherwise}.
    \end{cases}
\end{align}
This recursive expansion allows  MDP-Agent to iteratively refine and broaden the evidence pool, ensuring the resulting subspace \(\gK_i\) captures the full scope of knowledge necessary to resolve~the~intent.

\paragraph{Memory-Guided Parallel Synthesis.}
A second challenge is \emph{scalability}: as diffusion expands, the number of queries and retrieved documents grows rapidly, making exhaustive analysis prohibitively costly. To address this,  MDP-Agent employs \emph{memory-guided parallel evidence synthesis}, inspired by the map–reduce paradigm. Each retrieved document \(D\), equipped with its gist memory \(\gD\), is first processed by a filtering operator \(\mathcal{F}\) that performs lightweight relevance checks based solely on \(\gD\), discarding irrelevant candidates without accessing the full text. The surviving documents are then mapped in parallel by an extraction operator \(\mathcal{E}\) into fine-grained evidence units, which are subsequently reduced by a synthesis operator \(\mathcal{R}\) into the constructed subspace $\gK_i$ for intent \(I_i\):
\begin{align}
    \gK_i = \mathcal{R}\Big(\{\mathcal{E}(D) \;\mid\; D \in \mathcal{F}(\{D\}, I_i, \{\gD\})\}\Big).
\end{align}
All operators \(\mathcal{F}, \mathcal{E}, \mathcal{R}\) are powered by an auxiliary lightweight LLM, enabling  MDP-Agent to filter aggressively, extract in parallel, and synthesize compactly. This design ensures scalability and efficiency while preserving broad evidence coverage with manageable reasoning cost.

\paragraph{Task-Aware Contextualization.}
After evidence collection, MDP-Agent organizes the gathered information into a structured, task-specific context. Formally, given a task \(\gX\), the generated intents \(\{I_i\}\) and the constructed subspaces \(\{\gK_i\}\), the system assembles an organized knowledge chain:
\begin{align}
    \gC = \gX \;\cup\; (I_1 \rightarrow \gK_1) \;\rightarrow\; (I_2 \rightarrow \gK_2) \;\rightarrow \cdots \rightarrow (I_N \rightarrow \gK_N),
\end{align}
which explicitly encodes the reasoning trajectory and its supporting evidence. In essence, \(\gC\) represents the \emph{LLM-ready form} of the expected knowledge space \(\gK\): compact, structured, and directly consumable by an LLM. This task-specific context can then be fed into any downstream LLM to generate the final answer \(\gY\). We refer to this transformation from raw retrieval to structured, reasoning-ready context as MDP-Agent's \emph{contextual intelligence}.

In summary,  MDP-Agent is a retrieval framework that leverages an agentic paradigm to construct minimal yet sufficient LLM-ready knowledge representations for diverse tasks. Further implementation details are provided in Appendix~\ref{ap:imp}, and Table~\ref{tab:case} illustrates the processes through a case study.

\section{Experiments}
\subsection{Datasets and Baselines.}

\paragraph{Datasets.}  

We evaluate  MDP-Agent on two benchmarks for complex information-seeking.  
\textbf{GAIA} (General AI Assistant) comprises over 450 real-world queries spanning multi-step reasoning, multimodal understanding, and tool use~\citep{mialon2023gaia}. Following prior work~\citep{li2025webthinker,wu2025webdancer}, we use 103 text-only validation questions.  
\textbf{WebWalkerQA} includes 680 queries across domains such as conferences and organizations, requiring agents to traverse subpages and integrate dispersed evidence, which makes it a long-horizon reasoning challenge~\citep{wu2025webwalker}.  

\paragraph{Baselines.} 

We compare  MDP-Agent against three groups of baselines.  
(1) \emph{Direct Reasoning}: strong standalone LLMs used without external tools, including Qwen2.5-32B, Qwen2.5-32B, QwQ-32B, GPT-4o, Gemini-2.5-Flash and DeepSeek-R1-671B~\citep{deepseekr1,gemini25,openai2024gpt4technicalreport}.  
(2) \emph{Retrieval-Augmented Generation}: methods that inject retrieved evidence, such as vanilla RAG and enhanced variants with query planning or iterative refinement~\citep{ITER-RETGEN,chan2024rqraglearningrefinequeries}.  
(3) \emph{Tool-Integrated Reasoning}: approaches that interleave retrieval with reasoning, including ReAct, Search-o1, and WebThinker~\citep{yao2022react,searcho1,li2025webthinker}.  
Appendix~\ref{ap:imp} provides implementation details for  MDP-Agent and baselines.

\begin{table*}[t]
    \centering
    \small
    \caption{Main experimental results. Best scores are shown in bold, and second-best are \underline{underlined}. Following the official settings, we report Exact Match (EM) for GAIA, and LLM Equivalence Accuracy for WebWalkerQA.}    
\begin{tabular}{lp{.95cm}p{.95cm}p{.95cm}p{.6cm}p{.6cm}p{.8cm}p{.6cm}p{.6cm}}
\toprule
\multirow{2}[2]{*}{\textbf{Method}}& \multicolumn{4}{c}{\textbf{General AI Assistant}} & \multicolumn{4}{c}{\textbf{WebWalkerQA}}  \\
  & Level 1 & Level 2 & Level 3 & Avg. & Easy & Medium & Hard & Avg.\\

\midrule
\rowcolor{gray!12}\multicolumn{9}{c}{\textit{\textbf{Direct Reasoning (w/o Retrieval)}}} \\

Qwen2.5-32B & 20.5 & 9.6 & 8.3 & 13.6 & 3.8 & 2.5 & 3.3 & 3.1      \\
Qwen3-32B & 15.4 & 7.7 & 0.0 & 9.7 & 3.1 & 1.4 & 2.5 & 2.2\\
QwQ-32B   & 25.6 & 9.6 & 16.7 & 16.5  & 7.5 & 2.1 & 3.8 & 4.0 \\
GPT-4o  &   23.1 & 15.4 & 8.3 & 17.5 & 6.7 & 6.0 & 4.2 & 5.5    \\
Gemini-2.5-Flash & 33.3 & 11.5 & 0.0 & 18.5 & 16.3 & 7.9 & 5.8 & 9.1 \\
DeepSeek-R1-671B  & 43.6 & 26.9 & 8.3 & 31.1 & 5.0 & 11.8 & 11.3 & 10.0    \\
\rowcolor{gray!12}\multicolumn{9}{c}{\textit{\textbf{Retrieval-Augmented Generation}}} \\
Vanilla RAG (Qwen2.5-32B) & 12.8 & 11.8 & 8.3 & 11.8 & 23.1 & 14.3 & 11.3 & 15.3   \\
Vanilla RAG (QwQ-32B)    & 33.3 & 36.5 & 8.3 & 32.0 & 36.9 & 26.1 & 33.5 & 31.2  \\
Query Planning (Qwen2.5-32B)  & 30.8 & 17.3 & 0.0 & 20.4 & 29.4 & 36.4 & 25.0 & 30.7\\
Query Planning  (QwQ-32B) &  48.7 & 25.0  & 8.3 & 32.0 & 28.8 & 35.7 & 30.8 & 32.5\\

Iterative RAG  (Qwen2.5-32B)   &  35.9 &  19.2 &  8.3 & 24.3 & 30.6 & 35.7 & 25.4 & 30.9 \\
Iterative RAG (QwQ-32B)  & 51.3 & 28.8  & 8.3 & 35.0 & 29.4 & 32.9 & 31.3 &31.5 \\
\rowcolor{gray!12}\multicolumn{9}{c}{\textit{\textbf{Tool-Integrated Reasoning}}} \\

ReAct (Qwen2.5-32B)  & 46.1 &44.2& 8.3 & 40.7 & 44.3 & \underline{46.7} & 29.2 & 38.4 \\
ReAct (QwQ-32B) & 48.7 & 34.6 & 16.7 & 37.8 & 35.6 & 29.1 & 13.2 & 24.1 \\
ReAct (GPT-4o) & 51.2  & 34.6 & 8.3 & 34.6 & 34.6 & 42.0 &  23.9 & 33.8 \\
Search-o1-32B  & 53.8 &44.2 &16.7 & 39.8 & 43.1 & 35.0 & 27.1 & 34.1    \\

WebThinker-32B-Base & 53.8 &44.2 &16.7 & 44.7 & 47.5 & 41.1 & 39.2 & 41.9   \\
WebThinker-32B & 56.4 & \textbf{50.0} & 16.7 & \underline{48.5} & \textbf{58.8} & 44.6 & \underline{40.4} & \underline{46.5}\\
\midrule
\textbf{MDP-Agent}~(QwQ-32B) & \textbf{61.5} & \underline{46.2} & \textbf{33.3} & \textbf{50.5} & \underline{53.1} & \textbf{55.0} & \textbf{50.8} & \textbf{53.1} \\

\bottomrule
\end{tabular}
\label{tab:exp}
\end{table*}

\subsection{Main Results}
Table~\ref{tab:exp} reports the performance of  MDP-Agent and baseline.
Our key findings are as follows:  

(1) Under direct reasoning without retrieval, all models handle GAIA tasks more readily, yet their accuracy remains modest. By contrast, accuracy drops sharply on WebWalkerQA, confirming that these benchmarks demand recent and long-tail knowledge rarely captured in model parameters. Interestingly, Qwen3-32B, although more recent, underperforms both Qwen2.5-32B and QwQ-32B, suggesting that Qwen3’s hybrid reasoning design might compromise efficacy. Based on these observations, we use Qwen2.5-32B and QwQ-32B as the backbone models for RAG and agentic-search baselines.

(2)   MDP-Agent consistently outperforms not only vanilla RAG but also advanced variants that incorporate query rewriting or iterative refinement, validating the robustness of its retrieval paradigm. Unlike these pre-inference schemes that often leave evidence fragmented or incomplete,  MDP-Agent employs diffusion-based exploration and memory-guided synthesis to recover layered dependencies while filtering noise at scale. This design yields substantial gains on tasks that demand multi-hop reasoning and long-horizon synthesis, where RAG methods struggle to provide coherent context. 

(3)  MDP-Agent surpasses agentic-search baselines, including workflow-based methods (e.g., Search-o1) and end-to-end optimized systems (e.g., WebThinker). Although WebThinker benefits~from large-scale in-domain training,  MDP-Agent without task-specific optimization still outperforms WebThinker-Base across all dimensions and achieves dataset-level gains over WebThinker-32B-RL, falling only on two levels within the dataset hierarchy. This highlights MDP-Agent’s diffusive and parallel exploration, which enables broader coverage and reduces the risk of missing critical~evidence.

\subsection{Ablation Study}
\begin{figure}
    \centering
    \includegraphics[width=\linewidth]{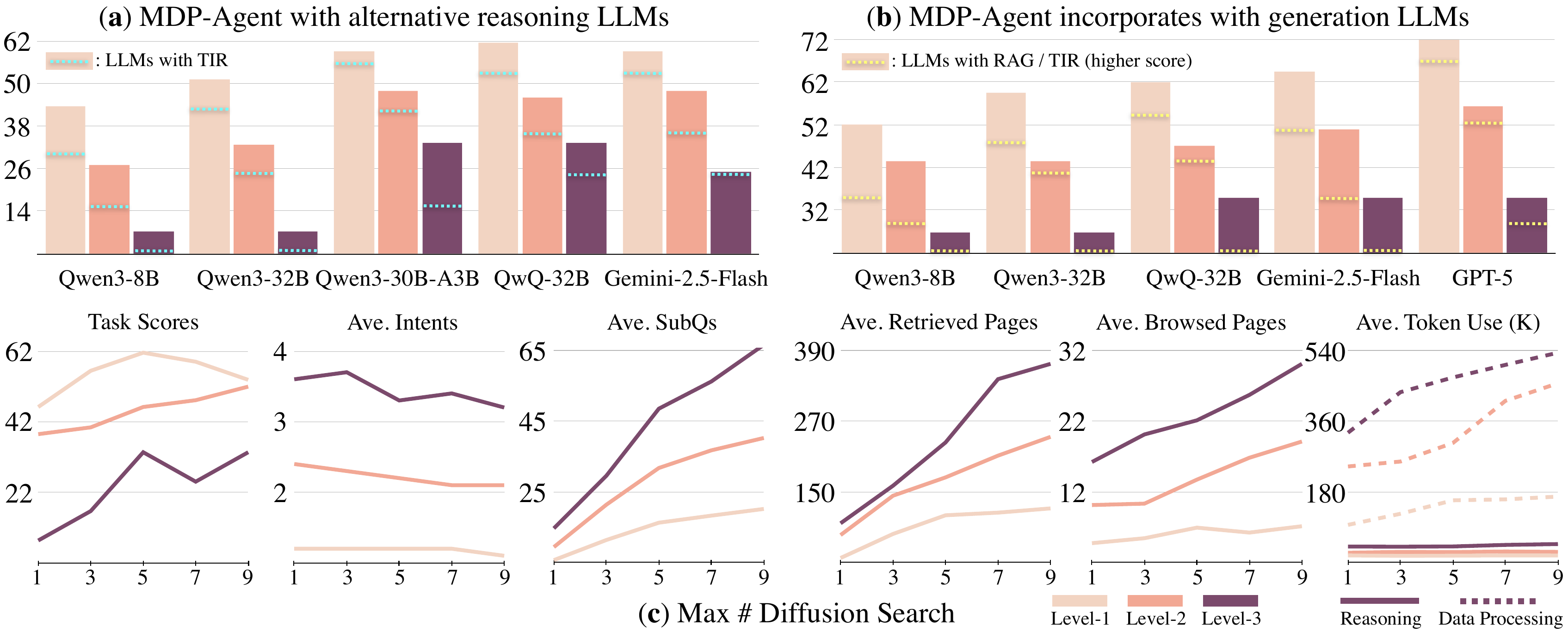}
    \vspace{-5pt}
    \caption{Analysis of  MDP-Agent on three perspectives: (a) effect of the central reasoning LLM, comparing  MDP-Agent with a TIR baseline (Search-o1); (b) transferability of MDP-Agent’s LLM-ready context, compared with RAG and TIR across downstream LLMs for answer generation; and (c)~impact of the diffusion-search budget on performance and resulting retrieval dynamics.
}
    \label{fig:abl}
\end{figure}

 MDP-Agent is designed as an integrated retrieval framework, operating as a unified system with interdependent components, making it more meaningful to analyze as a whole rather than in isolation.
Accordingly, our ablation study examines three dimensions: (1) the role of different LLMs as  MDP-Agent’s central reasoning agent, (2) the generalizability of  MDP-Agent-generated context across diverse models, and (3) the dynamics of agentic retrieval, with a focus on diffusion search depth and evidence synthesis efficiency. Figure~\ref{fig:abl} summarizes the results, which we discuss below.

\paragraph{Impact of Reasoning LLM Selection.}
 MDP-Agent relies critically on the capabilities of its central reasoning model. As shown in Figure~\ref{fig:abl}~(a), it consistently outperforms the tool-integrated reasoning baseline (Search-o1) across different LLMs, demonstrating the robustness of its design. Nonetheless, the strength of the reasoning model plays a decisive role. Reasoning-oriented models such as QwQ-32B and Qwen3-30B-A3B achieve the best results, clearly surpassing Qwen3-32B, a hybrid model with diluted reasoning capacity. When paired with Gemini2.5-Flash,  MDP-Agent also delivers competitive results through a dynamic strategy: the model enables “thinking mode” for complex planning steps while producing direct outputs for simpler ones, striking a balance between efficiency and accuracy. Overall, these findings show that while  MDP-Agent adapts well to diverse LLMs, its performance scales with the depth and quality of reasoning in the central agent.

\paragraph{Generalizability of MDP-Agent-Generated Context.}
 MDP-Agent serves as a retrieval framework that produces reliable, task-specific context in an LLM-ready form, which can be seamlessly applied to any model. As illustrated in Figure~\ref{fig:abl}~(b), supplying this curated context to different generation models consistently outperforms both TIR and RAG baselines, underscoring the robustness and generalizability of MDP-Agent. For smaller models such as Qwen3-8B, the performance gains are especially pronounced, showing that  MDP-Agent can effectively compensate for the limited reasoning and knowledge capacity of lightweight LLMs. Conversely, when applied to stronger models such as GPT-5, the curated context is leveraged even more effectively, yielding further improvements and demonstrating MDP-Agent’s scalability across model strengths.

\paragraph{Agentic Behavior across Diffusion Search Depths.}
Our analysis in Figure~\ref{fig:abl}~(c) highlights how MDP-Agent’s tailored techniques jointly contribute to effective and scalable retrieval. First, diffusion search proves critical: increasing its depth expands the evidence pool, improving task performance from depth 1 to 5 before fluctuating as information saturates. We also observe that deeper diffusion reduces the number of required search intents, easing the reasoning workload and accelerating convergence to answers. Second, deeper diffusion inevitably increases sub-queries and retrieved pages, especially for complex information-seeking tasks. Here, MDP-Agent’s memory-guided parallel evidence synthesis is validated: it filters out nearly 90\% of irrelevant pages using gist memory and processes the remainder in a map–reduce manner, demonstrating strong scalability. Finally, token usage analysis shows that reasoning accounts for a small fraction of total cost compared to large-scale data processing. This validates MDP-Agent’s design of assigning heavy reasoning to strong central models while outsourcing bulk data handling to lightweight auxiliary models, achieving an effective balance between efficiency and performance.

\subsection{Case Study}
Table~\ref{tab:case} presents a case study that illustrates how  MDP-Agent constructs an LLM-ready context for a complex task. The query requires first identifying a \emph{scientific genus} and then consulting academic papers, with the final answer derived by intersecting the animals discussed across these sources.

 MDP-Agent begins by leveraging the foundation model’s intrinsic knowledge to establish the target genus, which anchors subsequent information intents. For the first intent, retrieving the precise titles of the three papers,  MDP-Agent executes diffusive parallel exploration: issuing multiple atomic queries, gathering 36 candidate pages, filtering them with gist-level relevance checks, and browsing only 13 to distill the relevant evidence. This step demonstrates MDP-Agent’s ability to maximize coverage while keeping processing efficient. The process then advances to the next intent, shifting from paper discovery to full-text analysis, with new sub-queries generated adaptively to uncover the animals mentioned.
Once all necessary evidence is accumulated,  MDP-Agent synthesizes the results into a structured, task-specific context that integrates both retrieved knowledge and intermediate reasoning steps. The resulting representation is compact yet sufficient, capturing logical dependencies across intents, minimizing redundancy, and fitting neatly within the LLM’s context window. 

Notably, the case highlights MDP-Agent’s efficiency: reasoning consumes only 8.9K tokens, while large-scale evidence processing consumes 227K tokens. This demonstrates  MDP-Agent's balanced~design, where strong central models focus on reasoning while lightweight auxiliaries handle bulk text processing. As a result,  MDP-Agent achieves efficient exploration of massive raw text, effective evidence consolidation, and robust downstream reasoning without overloading the model with noise.

\begin{table*}[t]

\scriptsize
    \centering
    \caption{Case study on a Level-3 sample from GAIA. The reasoning agent within  MDP-Agent first addresses the initial knowledge gap using intrinsic knowledge, then conducts agentic external knowledge exploration. The resulting task-specific context reconstructs a minimal yet sufficient knowledge space and delivers it in an LLM-ready form for solving the input task.
}
    \begin{tabular}{p{.97\linewidth}}
    \toprule
     \textbf{Task}: What animals that were \textcolor{c3}{mentioned in both Ilias Lagkouvardos's and Olga Tapia's papers} on the alvei species of the genus named for Copenhagen outside the bibliographies were also \textcolor{c3}{present in the 2021 article cited on the alvei species' Wikipedia page} about a multicenter, randomized, double-blind study?
     \quad  \textbf{Ground-truth answer}:  \textcolor{c3}{Mice}\\
\midrule

\textbf{Initial Reasoning}: Identify the scientific genus named for Copenhagen. $\rightarrow$ \textcolor{c1}{Hafnia}

 \textbf{Information Intent $\gI_1$}: \\

- Find scientific papers by Ilias Lagkouvardos concerning Hafnia alvei.

- Find scientific papers by Olga Tapia concerning Hafnia alvei.

- Find the Wikipedia page for Hafnia alvei to locate the 2021 article cited about a multicenter, randomized, double-blind study.

 \textcolor{c2}{\textbf{\textit{4 atomic queries issued → 36 pages retrieved → 13 pages browsed; Tokens: 8.9K (reasoning), 227.2K (processing).}}} \\
\textbf{Knowledge Subspace $\gK_1$}: Ilias Lagkouvardos authored a paper ``\textcolor{c1}{An Insight into Goat Cheese: The Tales of Artisanal and Industrial Gidotyri Microbiota}'' on H. alvei were identified, while Olga Tapia authored a 2022 paper in Molecules titled “\textcolor{c1}{The Molecular Weaponry Produced by the Bacterium Hafnia alvei in Foods}”, discussing the bacterium’s role in food systems. The Wikipedia page for Hafnia alvei cites a 2021 multicenter, randomized, double-blind study published in Nutrients—``\textcolor{c1}{The probiotic strain H. alvei HA4597® improves weight loss in overweight subjects under moderate hypocaloric diet}''.

\textbf{Information Intent $\gI_2$}: \\
- Search for the full text of Olga Tapia's 2022 paper ``The Molecular Weaponry Produced by the Bacterium Hafnia alvei in Foods'' published in Molecules and identify any animals mentioned within the text (excluding bibliographies).
...... [omitted]\\
\midrule

   \textbf{Task-Specific Context $\gC$}: 
The task requires the following steps and information:

 \textcolor{c1}{Step 1}: Identify the scientific genus named for Copenhagen and confirm the relevant alvei species.
 \textcolor{c1}{Info}: The genus is Hafnia, derived from the Latin name for Copenhagen; relevant species: Hafnia alvei.

 \textcolor{c1}{Step 2}: Identify the 2021 article cited on the Hafnia alvei Wikipedia page.
 \textcolor{c1}{Info}: Nutrients (2021): “The probiotic strain H. alvei HA4597® improves weight loss in overweight subjects under moderate hypocaloric diet” (multicenter, randomized, double-blind).

...... [omitted]

 \textcolor{c1}{Step 5}: Identify animals mentioned in the 2021 Nutrients study.
 \textcolor{c1}{Info}: Human participants (n=236); also referenced preclinical studies using obese mice (ob/ob, HFD-fed).

 \textcolor{c1}{Step 6}: Determine overlap of animals across sources.
 \textcolor{c1}{Info}: The only animal shared in the papers is mice. \\

    \bottomrule
    \end{tabular}
    \label{tab:case}

\end{table*}

\section{Conclusion}

In this work, we introduced the \textbf{Model–Document Protocol (MDP)}, a general framework that formalizes how unstructured documents are transformed into structured, task-specific, and LLM-ready knowledge representations. MDP redefines the retrieval–reasoning interface by establishing principled pathways that convert \emph{data chaos} into \emph{knowledge order}, ensuring that what reaches the model is not raw fragments but coherent, verifiable context.

As an instantiation of this protocol, we presented \textbf{MDP-Agent}, a retrieval framework that iteratively explores and exploits external knowledge sources. At the corpus level, MDP-Agent abstracts large unstructured corpora into document-level gist memories, achieving global semantic coverage while encoding implicit structure often missed by conventional dense retrieval. At the task level, it engages in an agentic reasoning process that decomposes complex queries into layered intents, performs diffusion-based exploration with vertical exploitation, and synthesizes the resulting evidence into compact, LLM-consumable context. Through this process, MDP-Agent reconstructs the minimal yet sufficient knowledge space required to close task-specific knowledge gaps.

Extensive experiments on diverse information-seeking benchmarks, supported by ablation analyses and detailed case studies, validate both the soundness of the MDP framework and the effectiveness of its agentic instantiation. Together, these results demonstrate that MDP and MDP-Agent provide a principled, scalable, and general paradigm for bridging raw documents and reasoning-ready knowledge, empowering large language models with genuine contextual intelligence.

\bibliography{iclr2026_conference}

\begin{thebibliography}{22}
\providecommand{\natexlab}[1]{#1}
\providecommand{\url}[1]{\texttt{#1}}
\expandafter\ifx\csname urlstyle\endcsname\relax
  \providecommand{\doi}[1]{doi: #1}\else
  \providecommand{\doi}{doi: \begingroup \urlstyle{rm}\Url}\fi

\bibitem[Chan et~al.(2024)Chan, Xu, Yuan, Luo, Xue, Guo, and Fu]{chan2024rqraglearningrefinequeries}
Chi{-}Min Chan, Chunpu Xu, Ruibin Yuan, Hongyin Luo, Wei Xue, Yike Guo, and Jie Fu.
\newblock {RQ-RAG:} learning to refine queries for retrieval augmented generation.
\newblock \emph{CoRR}, abs/2404.00610, 2024.
\newblock \doi{10.48550/ARXIV.2404.00610}.

\bibitem[Chen et~al.(2023)Chen, Xiao, Zhang, Luo, Lian, and Liu]{bge_m3}
Jianlv Chen, Shitao Xiao, Peitian Zhang, Kun Luo, Defu Lian, and Zheng Liu.
\newblock Bge m3-embedding: Multi-lingual, multi-functionality, multi-granularity text embeddings through self-knowledge distillation, 2023.

\bibitem[DeepSeek{-}AI(2025)]{deepseekr1}
DeepSeek{-}AI.
\newblock Deepseek-r1: Incentivizing reasoning capability in llms via reinforcement learning.
\newblock \emph{CoRR}, abs/2501.12948, 2025.
\newblock \doi{10.48550/ARXIV.2501.12948}.

\bibitem[Gao et~al.(2024)Gao, Xiong, Gao, Jia, Pan, Bi, Dai, Sun, Guo, Wang, and Wang]{gao2024retrievalaugmented}
Yunfan Gao, Yun Xiong, Xinyu Gao, Kangxiang Jia, Jinliu Pan, Yuxi Bi, Yi~Dai, Jiawei Sun, Qianyu Guo, Meng Wang, and Haofen Wang.
\newblock Retrieval-augmented generation for large language models: A survey, 2024.

\bibitem[Gemini~Team(2025)]{gemini25}
Google Gemini~Team.
\newblock Gemini 2.5: Pushing the frontier with advanced reasoning, multimodality, long context, and next generation agentic capabilities, 2025.

\bibitem[Huang et~al.(2025)Huang, Yu, Ma, Zhong, Feng, Wang, Chen, Peng, Feng, Qin, et~al.]{hallucination}
Lei Huang, Weijiang Yu, Weitao Ma, Weihong Zhong, Zhangyin Feng, Haotian Wang, Qianglong Chen, Weihua Peng, Xiaocheng Feng, Bing Qin, et~al.
\newblock A survey on hallucination in large language models: Principles, taxonomy, challenges, and open questions.
\newblock \emph{ACM Transactions on Information Systems}, 43\penalty0 (2):\penalty0 1--55, 2025.

\bibitem[Jin et~al.(2025)Jin, Zeng, Yue, Wang, Zamani, and Han]{searchr1}
Bowen Jin, Hansi Zeng, Zhenrui Yue, Dong Wang, Hamed Zamani, and Jiawei Han.
\newblock Search-r1: Training llms to reason and leverage search engines with reinforcement learning.
\newblock \emph{CoRR}, abs/2503.09516, 2025.
\newblock \doi{10.48550/ARXIV.2503.09516}.

\bibitem[Lewis et~al.(2020)Lewis, Perez, Piktus, Petroni, Karpukhin, Goyal, K{\"u}ttler, Lewis, Yih, Rockt{\"a}schel, et~al.]{rag20}
Patrick Lewis, Ethan Perez, Aleksandra Piktus, Fabio Petroni, Vladimir Karpukhin, Naman Goyal, Heinrich K{\"u}ttler, Mike Lewis, Wen-tau Yih, Tim Rockt{\"a}schel, et~al.
\newblock Retrieval-augmented generation for knowledge-intensive nlp tasks.
\newblock \emph{Advances in Neural Information Processing Systems}, 33:\penalty0 9459--9474, 2020.

\bibitem[Li et~al.(2025{\natexlab{a}})Li, Dong, Jin, Zhang, Zhou, Zhu, Zhang, and Dou]{searcho1}
Xiaoxi Li, Guanting Dong, Jiajie Jin, Yuyao Zhang, Yujia Zhou, Yutao Zhu, Peitian Zhang, and Zhicheng Dou.
\newblock Search-o1: Agentic search-enhanced large reasoning models.
\newblock \emph{CoRR}, abs/2501.05366, 2025{\natexlab{a}}.
\newblock \doi{10.48550/ARXIV.2501.05366}.

\bibitem[Li et~al.(2025{\natexlab{b}})Li, Jin, Dong, Qian, Zhu, Wu, Wen, and Dou]{li2025webthinker}
Xiaoxi Li, Jiajie Jin, Guanting Dong, Hongjin Qian, Yutao Zhu, Yongkang Wu, Ji-Rong Wen, and Zhicheng Dou.
\newblock Webthinker: Empowering large reasoning models with deep research capability.
\newblock \emph{arXiv preprint arXiv:2504.21776}, 2025{\natexlab{b}}.

\bibitem[Mialon et~al.(2023)Mialon, Fourrier, Wolf, LeCun, and Scialom]{mialon2023gaia}
Gr{\'e}goire Mialon, Cl{\'e}mentine Fourrier, Thomas Wolf, Yann LeCun, and Thomas Scialom.
\newblock Gaia: a benchmark for general ai assistants.
\newblock In \emph{The Twelfth International Conference on Learning Representations}, 2023.

\bibitem[OpenAI(2024)]{openai2024gpt4technicalreport}
OpenAI.
\newblock Gpt-4 technical report, 2024.

\bibitem[Ouyang et~al.(2022)Ouyang, Wu, Jiang, Almeida, Wainwright, Mishkin, Zhang, Agarwal, Slama, Ray, et~al.]{chatgpt}
Long Ouyang, Jeffrey Wu, Xu~Jiang, Diogo Almeida, Carroll Wainwright, Pamela Mishkin, Chong Zhang, Sandhini Agarwal, Katarina Slama, Alex Ray, et~al.
\newblock Training language models to follow instructions with human feedback.
\newblock \emph{Advances in neural information processing systems}, 35:\penalty0 27730--27744, 2022.

\bibitem[Qian et~al.(2023)Qian, Zhu, Dou, Gu, Zhang, Liu, Lai, Cao, Nie, and Wen]{qian2023webbrain}
Hongjing Qian, Yutao Zhu, Zhicheng Dou, Haoqi Gu, Xinyu Zhang, Zheng Liu, Ruofei Lai, Zhao Cao, Jian-Yun Nie, and Ji-Rong Wen.
\newblock Webbrain: Learning to generate factually correct articles for queries by grounding on large web corpus, 2023.

\bibitem[Shao et~al.(2023)Shao, Gong, Shen, Huang, Duan, and Chen]{ITER-RETGEN}
Zhihong Shao, Yeyun Gong, Yelong Shen, Minlie Huang, Nan Duan, and Weizhu Chen.
\newblock Enhancing retrieval-augmented large language models with iterative retrieval-generation synergy.
\newblock \emph{arXiv preprint arXiv:2305.15294}, 2023.

\bibitem[Wu et~al.(2025{\natexlab{a}})Wu, Li, Fang, Yin, Zhang, Tao, Zhang, Xi, Fu, Jiang, et~al.]{wu2025webdancer}
Jialong Wu, Baixuan Li, Runnan Fang, Wenbiao Yin, Liwen Zhang, Zhengwei Tao, Dingchu Zhang, Zekun Xi, Gang Fu, Yong Jiang, et~al.
\newblock Webdancer: Towards autonomous information seeking agency.
\newblock \emph{arXiv preprint arXiv:2505.22648}, 2025{\natexlab{a}}.

\bibitem[Wu et~al.(2025{\natexlab{b}})Wu, Yin, Jiang, Wang, Xi, Fang, Zhang, He, Zhou, Xie, et~al.]{wu2025webwalker}
Jialong Wu, Wenbiao Yin, Yong Jiang, Zhenglin Wang, Zekun Xi, Runnan Fang, Linhai Zhang, Yulan He, Deyu Zhou, Pengjun Xie, et~al.
\newblock Webwalker: Benchmarking llms in web traversal.
\newblock \emph{arXiv preprint arXiv:2501.07572}, 2025{\natexlab{b}}.

\bibitem[Yao et~al.(2023)Yao, Zhao, Yu, Du, Shafran, Narasimhan, and Cao]{yao2022react}
Shunyu Yao, Jeffrey Zhao, Dian Yu, Nan Du, Izhak Shafran, Karthik Narasimhan, and Yuan Cao.
\newblock React: Synergizing reasoning and acting in language models.
\newblock In \emph{International Conference on Learning Representations (ICLR)}, 2023.

\bibitem[Zhang et~al.(2024)Zhang, Liao, Li, Du, and Lin]{zhang2024agentic}
Weinan Zhang, Junwei Liao, Ning Li, Kounianhua Du, and Jianghao Lin.
\newblock Agentic information retrieval.
\newblock \emph{arXiv preprint arXiv:2410.09713}, 2024.

\bibitem[Zhao et~al.(2024{\natexlab{a}})Zhao, Yang, Wang, He, Qiu, and Qiu]{zhao2024retrievalaugmentedgenerationrag}
Siyun Zhao, Yuqing Yang, Zilong Wang, Zhiyuan He, Luna~K. Qiu, and Lili Qiu.
\newblock Retrieval augmented generation (rag) and beyond: A comprehensive survey on how to make your llms use external data more wisely, 2024{\natexlab{a}}.

\bibitem[Zhao et~al.(2024{\natexlab{b}})Zhao, Zhou, Li, Tang, Wang, Hou, Min, Zhang, Zhang, Dong, Du, Yang, Chen, Chen, Jiang, Ren, Li, Tang, Liu, Liu, Nie, and Wen]{zhao2024surveylargelanguagemodels}
Wayne~Xin Zhao, Kun Zhou, Junyi Li, Tianyi Tang, Xiaolei Wang, Yupeng Hou, Yingqian Min, Beichen Zhang, Junjie Zhang, Zican Dong, Yifan Du, Chen Yang, Yushuo Chen, Zhipeng Chen, Jinhao Jiang, Ruiyang Ren, Yifan Li, Xinyu Tang, Zikang Liu, Peiyu Liu, Jian-Yun Nie, and Ji-Rong Wen.
\newblock A survey of large language models, 2024{\natexlab{b}}.

\bibitem[Zhu et~al.(2024)Zhu, Yuan, Wang, Liu, Liu, Deng, Chen, Dou, and Wen]{zhu2024largelanguagemodelsinformation}
Yutao Zhu, Huaying Yuan, Shuting Wang, Jiongnan Liu, Wenhan Liu, Chenlong Deng, Haonan Chen, Zhicheng Dou, and Ji-Rong Wen.
\newblock Large language models for information retrieval: A survey, 2024.

\end{thebibliography}
\bibliographystyle{iclr2026_conference}

\appendix
\section{Implementation Details}
\label{ap:imp}

In the main experiments,  MDP-Agent adopts QwQ-32B as the central reasoning model, supported by Qwen3-7B as an auxiliary processor for parallel data synthesis. For each query,  MDP-Agent curates a task-specific context, which is then directly fed into standalone LLMs for answer generation. Unless specified, the maximum diffusion depth for the \emph{diffusive wide exploration} is set to 5.  

To construct the web page collection used in the benchmarks, we first run  MDP-Agent directly with a search engine. In this initialization step, the local index is replaced by the search engine, and the use of gist memory is approximated by the first 1,024 tokens of each retrieved page. For every query, the top-20 web pages are collected. After five full runs on each benchmark, this procedure yields approximately 100K web pages in total. These pages are then processed to generate gist memories, which serve as the basis for the subsequent indexing process within MDP-Agent.  

Notably, evaluation with the online search engine proves both \emph{slow} and \emph{unstable}, since each sample requires executing multiple sub-queries and crawling tens to hundreds of web pages. The performance in this setting is also substantially lower than that achieved with the local index.  

During the \emph{data indexing process}, we employ BGE-M3 as the dense embedding model~\citep{bge_m3}, complemented by a BM25 index constructed over the full web content. All retrieval operations are instantiated using ElasticSearch, which provides a stable and scalable infrastructure for large-scale search.  
For online retrieval, we rely on Google’s \href{https://developers.google.com/custom-search/v1/overview}{Custom Search JSON API} to identify relevant pages, and utilize Jina AI’s \href{https://jina.ai/api-dashboard/reader}{Web Reader} to extract full web content.  
For all baselines, we either report results directly from their original papers or reproduce them using official implementations. All experiments are conducted on a node of eight NVIDIA A100-40G GPUs.  

To ensure transparency and reproducibility, we release all prompts used in  MDP-Agent along with full experiment logs, including intermediate artifacts such as search intents, atomic queries, retrieved and browsed pages, refined evidence, and organized knowledge. 

\end{document}